\documentclass[a4paper,fleqn]{cas-dc}

\usepackage{natbib}
\setcitestyle{authoryear,open={(},close={)}} %Citation-related commands

\usepackage{multirow}
\usepackage[utf8]{inputenc}
\usepackage{array,makecell}% <-- added makecell
\usepackage{booktabs}
\def\tsc#1{\csdef{#1}{\textsc{\lowercase{#1}}\xspace}}
\tsc{WGM}
\tsc{QE}
\tsc{EP}
\tsc{PMS}
\tsc{BEC}
\tsc{DE}
\begin{document}
\let\WriteBookmarks\relax
\def\floatpagepagefraction{1}
\def\textpagefraction{.001}

\shorttitle{Scalable water mapping with airborne LiDAR data}
% \shorttitle{Scalable water mapping with point clouds}

\shortauthors{H. Song et~al.}

\title [mode = title]{A Fully Automated and Scalable Surface Water Mapping with Topographic Airborne LiDAR Data}                      
% \tnotemark[1,2]

% % Title footnote 1.
% % eg: \tnotetext[1]{Title footnote text}
% % \tnotetext[<tnote number>]{<tnote text>} 
% \tnotetext[1]{This document is the results of the research
%    project funded by the National Science Foundation.}

% \tnotetext[2]{The second title footnote which is a longer text matter
%    to fill through the whole text width and overflow into
%    another line in the footnotes area of the first page.}

% \author[1]{Hunsoo Song}
\author[1]{Hunsoo Song}[orcid=0000-0001-6899-6770]
% [type=editor,
%                         auid=000,bioid=1,
%                         % prefix=Sir,
%                         role=Researcher,
%                         orcid=0000-0001-7511-2910]

% \cormark[2]


\ead{hunsoo@purdue.edu}

% % URL of the first author
% \ead[url]{www.cvr.cc, cvr@sayahna.org}

% %  Credit authorship
% \credit{Conceptualization of this study, Methodology, Software}

% \affiliation[1]{organization={The Lyles School of Civil Engineering, Purdue University},
%     addressline={550 Stadium Mall Drive}, 
%     city={West Lafayette},
%     % citysep={}, % Uncomment if no comma needed between city and postcode
%     postcode={IN 47907}, 
%     % state={},
%     country={the United States}}
\address[1]{The Lyles School of Civil Engineering, Purdue University, 550 Stadium Mall Drive, West Lafayette, Indiana, the United States}
\author[1]{ Jinha Jung}[orcid=0000-0003-1176-3540]%[style=chinese]
\cormark[1]
% % Third author
% \author[2,3]{CV Rajagopal}[%
%    role=Co-ordinator,
%    suffix=Jr,
\ead{jinha@purdue.edu}
% \ead[URL]{www.sayahna.org}

% \credit{Data curation, Writing - Original draft preparation}

% % Address/affiliation
% \affiliation[2]{organizpreprintation={Sayahna Foundation},
%     % addressline={}, 
%     city={Jagathy},
%     % citysep={}, % Uncomment if no comma needed between city and postcode
%     postcode={695014}, 
%     state={Trivandrum},
%     country={India}}

% % Fourth author
% \author%
% [1,3]
% {Rishi T.}
% \cormark[2]
% \fnmark[1,3]
% \ead{rishi@stmdocs.in}
% \ead[URL]{www.stmdocs.in}

% \affiliation[3]{organization={STM Document Engineering Pvt Ltd.},
%     addressline={Mepukada}, 
%     city={Malayinkil},
%     % citysep={}, % Uncomment if no comma needed between city and postcode
%     postcode={695571}, 
%     state={Trivandrum},
%     country={India}}

\cortext[cor1]{Corresponding author}
% \cortext[cor2]{Principal corresponding author}

% \fntext[fn1]{This is the first author footnote. but is common to third
% \fntext[fn2]{Another author footnote, this is a very long footnote and
%   it should be a really long footnote. But this footnote is not yet
%   sufficiently long enough to make two lines of footnote text.}

% % For a title note without a number/mark
% \nonumnote{This note has no numbers. In this work we demonstrate $a_b$
%   the formation Y\_1 of a new type of polariton on the interface
%   between a cuprous oxide slab and a polystyrene micro-sphere placed

\begin{abstract}
Reliable and accurate high-resolution maps of surface waters are critical inputs to models that help understand the impacts and relationships between the environment and human activities. Advances in remote sensing technology have opened up the possibility of mapping very small bodies of water that are closely related to people's daily lives and are mostly affected by anthropogenic pressures. However, a robust and scalable method that works well for all types of water bodies located in diverse landscapes at high-resolution has yet to be developed.
This paper presents a method that can accurately extract surface water bodies up to a very fine scale in a wide variety of landscapes. Unlike optical image-based methods, the proposed method exploits the robust assumption that surface water is flat as gravity always pulls liquid molecules down. Based on this natural law, the proposed method extracts accurate, high-resolution water bodies including their elevations in a fully automated manner using only airborne LiDAR data. Extensive experiments with large ($\approx$ 2,500$km^{2}$) and diverse landscapes (urban, coastal, and mountainous areas) confirmed that our method can generate accurate results without site-specific parameter tunings for varied types of surface water. 
The proposed method enables an automated, scalable high-resolution mapping of a full 3D topography that includes both water and terrain, using only point clouds for the first time. We will release the code to the public in the hope that our work would lead to more effective solutions to help build a sustainable and resilient environment.

% \noindent\texttt{\textbackslash begin{abstract}} \dots 
% \texttt{\textbackslash end{abstract}} and
% \verb+\begin{keyword}+ \verb+...+ \verb+\end{keyword}+ 
% which
% contain the abstract and keywords respectively. 

% \noindent Each keyword shall be separated by a \verb+\sep+ command.
\end{abstract}

% \includegraphics{figs/grabs.pdf}

\begin{highlights}
\item We present an automated and scalable surface water mapping method.
\item Our method extracts surface water based on the natural law that surface water is flat.
\item Our method is particularly advantageous in steep mountains and urban areas with very small water bodies.
\item Our method enables a full 3D topography mapping using only airborne LiDAR data. 
\end{highlights}

% Keywords
\begin{keywords}
Surface water mapping \sep Airborne LiDAR data \sep Region merging \sep Small water bodies
\end{keywords}

\maketitle

\section{Introduction}
Surface water maps provide important knowledge to help understand the impacts and relationships between the environment and human activities. Knowing where water is and how it changes and relates to the surrounding environment enables successful responses that are closely related to water resources \citep{poff2016sustainable,viviroli2011climate,florke2018water}, biodiversity \citep{cawse2021nasa}, and human well-being \citep{sanders2022large}.

Optical remote sensing technology has become the most widely adopted and powerful tool for surface water mapping \citep{huang2018detecting}. Standard practice for extracting water using optical imaging is to exploit the spectral characteristics of water that the reflectance of water is significantly lower in the infrared channel than other channels compared to other objects \citep{xu2006modification, ji2009analysis}. Large bodies of water, such as oceans, rivers, and lakes, have been the focus of water mappings since early Earth observation satellite missions \citep{pekel2016high,khandelwal2017approach}, and reflectance-based water mapping has shown successful performance on medium- and low-resolution ($>$100$m$) satellite imagery \citep{feyisa2014automated,yamazaki2015development}. 

With the development of high-resolution imaging technology, mapping of very small water bodies ($<$1$ha$ or 10,000$m^2$) also become possible. Particularly, as small water bodies, such as small lakes, ponds, low-level streams, and ditches, are closely related to people's daily lives and are highly affected by anthropogenic footprints, the demand for a very high-resolution map of water bodies has been rapidly increasing these days \citep{kelly2022opportunities,xu2020towards,hoekstra2018urban,biggs2017importance,riley2018small}. However, in high-resolution images, conventional reflectance-based water mapping does not show satisfactory performance \citep{ogilvie2018surface} and often underestimates the extent of small water bodies \citep{pickens2020mapping,mao2022high}. This is mainly because the reflectance of small water bodies is highly varying in high-resolution images while high-resolution images require multiple image acquisitions to cover large-area which will again significantly increase the spectral variability. 

% As a result, reflectance-based methods often do not produce satisfactory results for high-resolution, large-area waterbody mapping. 
% Urban surface water body detection with suppressed built-up noise based on water indices from Sentinel-2 MSI imagery (Yang2018)

% To address this problem, efforts have been made to develop more sophisticated algorithms. Integrating data of multi-temporal \citep{pickens2020mapping} or different types of sensors (e.g., radar)\citep{ahmad2019fusion} and developing data-driven methods \citep{ko2015classification,sun2015soft, chen2020novel,isikdogan2017surface, wang2020urban} are examples. However, these approaches also have a problem in that it is difficult to consistently derive satisfactory results for multiple images acquired under various conditions. 

To address this problem, efforts have been made to develop more sophisticated methods. Integrating multi-temporal \citep{pickens2020mapping} or different types of sensors (e.g., radar) \citep{ahmad2019fusion} or developing data-driven methods \citep{ko2015classification,isikdogan2017surface, chen2020novel, wang2020urban} are examples. 
However, fundamentally, algorithms based on optical images cannot be free from spectral characteristics of water whose values vary depending on image acquisition conditions e.g., atmospheric conditions and sensor specifications) \citep{martins2017assessment} or surface water conditions (e.g., turbidity and surface roughness) \citep{odermatt2012review,kim2016evaluation}.
%However, fundamentally, algorithms based on optical images are not free from spectral characteristics of water whose value varies according to either the image acquisition conditions (e.g., atmospheric conditions and sensor specifications) \citep{martins2017assessment} or the varying surface conditions of water (e.g., turbidity and surface roughness) \citep{odermatt2012review,kim2016evaluation}. In other words, a robust and scalable method that works well for all water bodies located in a variety of landscapes has yet to be developed.

In addition, current optical image-based water mapping algorithms are limited to providing only 2D information. Meanwhile, even simply for identifying the direction of water flow, 3D topographic information is needed. Obviously, 3D water information can provide better information for understanding the dynamics of water, and it enables in-depth analysis of how water relates to its surrounding environments and improves the management of water resources and water-related disasters \citep{musa2015review,jarihani2015satellite,arrighi2019effects}. 
Thus, the authors note that airborne laser scanning will eventually be required for surface water maps to be truly useful as the LiDAR data is needed to obtain high-resolution 3D topographic information. Nevertheless, as airborne laser scanning has not been effectively used for surface water mapping, separate resources and efforts have been required for completing a 3D hydrography. 
%In response to the growing demand for 3D water information, the United States Geological Survey recently established the 3D Hydrography Program (3DHP) initiative to renew the Nation’s hydrography data (https://www.usgs.gov/national-hydrography/3d-national-topography-model-call-action-part-1-3d-hydrography-program).

This paper presents a scalable water mapping method using topographic airborne LiDAR data. The scalable method refers to the method that yields consistently satisfactory results across wide and varied landscapes. Our method operates in a fully automated manner, only requires topographic airborne LiDAR data, and extracts accurate and high-resolution water bodies including the elevation of water bodies. Extensive experiments with large ($\approx$ 2,500$km^{2}$) and diverse landscapes (urban, coastal, and mountainous areas) confirmed that our algorithm can generate more accurate results than NDWI-based methods even without any additional parameter tunings. Moreover, the proposed method enables high-resolution mapping of a full 3D topographic map (water and terrain) with only airborne LiDAR data. %Also, as a beneficial by-product, a 3D water map is also automatically generated.

The remainder of this paper is organized as follows. Section 2 illustrates related works. Section 3 describes the proposed surface water mapping methods. Section 4 illustrates datasets and experimental design. Section 5 provides experimental results. Section 6 concludes the paper.

\section{Related works}
\subsection{Optical imagery-based water mapping}

Optical imagery has been dominantly used for surface water mapping. With multiple satellite missions at different spatial resolutions ranging from 1 km to 1-m resolution, water bodies of the Earth have been mapped for more than several decades \citep{pekel2016high,khandelwal2017approach}. In particular, with recent advancements in satellite, airborne, and UAV imaging technology, a large-area mapping of small water bodies has attracted more attention \citep{becker2019unmanned, straffelini2021mapping,qayyum2020glacial, sogno2022remote}.

Although the higher spatial resolution imagery enables observing small water bodies, large-area mapping of small water bodies is more complicated than that with the lower resolution imageries for large water bodies. This is because, first, small water bodies have more wide variety of spectral reflectance \citep{ogilvie2018surface}. Second, shadow, occlusion, and relief displacement of urban structures make water mapping more challenging \citep{yang2018urban}
. Third, large-area mapping with high-resolution imagery necessitates multiple different image acquisitions, while different acquisitions may require different threshold values for water index-based mapping \citep{feyisa2014automated}.

%instead, we totally exclude optical characteristics of water for the first time, combatin spectral variability especially in high resolution 
To improve the accuracy, some recent studies have developed more sophisticated decision rules. 
\citep{yang2018urban} alleviated errors from built-up areas by utilizing multiple water and shadow indices. 
% \citep{yang2018urban} alleviated errors from built-up areas by utilizing the difference between the MNDWI (modified normalized difference water index) and AWEI (automated water extraction index) and integrating the shadow index. 
% To address the instability of spectral information,
\citep{dong2022mapping} integrated the spatial information into the classification rule by introducing the roughness of the water index. 
% The algorithm first identifies potential water and certainly water classes based on the water index and refines results based on the spatial connectivity and the water index roughness, which describes the spectral variability within neighboring pixels.
In addition, there have been studies that used data-driven approaches such as support vector machine \citep{sun2015soft}, random forest \citep{ko2015classification}, and deep learning methods \citep{chen2020novel, isikdogan2017surface, wang2020urban}.
% In addition, there have been studies that used data-driven approaches for water mappings. Machine learning algorithms, such as support vector machine \citep{sun2015soft}, random forest \citep{ko2015classification}, and deep learning \citep{chen2020novel, isikdogan2017surface, wang2020urban}, were successfully used for surface water mapping. 

More often, water has been mapped with other multiple classes including ice and cloud \citep{wu2018scribble} or more comprehensive land covers \citep{song2019patch, robinson2019large,zhang2022urbanwatch}. Most methods obtained satisfactory results under the condition that sufficient high-quality training samples are available. 
However, acquiring high-quality training samples for large areas is a challenging task, particularly in high-resolution image, for the same reason that surface water mapping using spectral characteristics is difficult \cite{pickens2020mapping, liu2021production,murray2022high}.
%However, acquiring high-quality training samples in large-area mapping with high-resolution images are challenging, and it is susceptible to training sample biased problems and easily faces out-of-distribution problem. 
Moreover, the decision rule of data-driven approaches is generally not clear and their results are biased towards the training data distribution \citep{moreno2012unifying}, which makes quality control of their output difficult.
% Moreover, as the decision boundary of data driven approaches is not definitive, errors are not generally explainable, which can pose a quality control issue as their uncertainties are not yet fully discovered. 
% This is why most of large-area water body mapping in practice employ a water-index based mapping \citep{huang2018detecting, xu2020towards}.

\subsection{Airborne LiDAR based water mapping}

Relatively very few studies have used topography airborne laser scanning for surface water mapping. Topographic airborne laser scanning refers to the system operated with the near infrared laser. We exclude bathymetry LiDAR in our discussion. The main idea of surface water mapping with LiDAR is to take advantage of the intensity values of airborne LiDAR and the property that the point density of LiDAR point clouds within water areas is much lower than in non-water areas as water hardly reflects the laser point.

%\citep{brzank2005aspects} classified low-point density areas as water bodies and performed a region-growing-based water mapping using the height and intensity values. 
%\citep{brzank2006classification} = preliminary of \citep{brzank2008aspects}
\citep{brzank2008aspects} performed a supervised fuzzy classification using using height, intensity, and point density as input. \citep{hofle2009water} modeled both intensity and point density based on the sensing geometry and performed a seeded region growing segmentation and object-based classification for surface water mapping. \citep{smeeckaert2013large} used a support vector machine for water extraction based on historical coastline data. This study refined the coastline data and conducted a supervised classification using 3D point-based features. \citep{crasto2015lidar} performed a decision tree-based data-driven approach using point density, elevation, and intensity of LiDAR points as inputs. More recently, \citep{malinowski2016local} input radiometric and geometric laser point features of full-waveform LiDAR into supervised classifiers to conduct flood mapping. \citep{shaker2019automatic} input multispectral LiDAR-based features into the log-likelihood classification model and performed supervised classification for water mapping. Also, \citep{yan2019scan} performed water body mapping by developing a method that collects training samples based on the ratio of intensity and elevation for each scan line and performs a classification using the maximum likelihood model.

% Also, (Yan 2019) developed a method for water body mapping using the ratio of intensity and elevation for each scan line, and (Yan 2021) further improved its performance and generated hydro-flattened DEM by improving the laser dropout modeling. % Also, (Yan 2019) utilized the ratio of intensity and elevation for each scan line for water body mapping, and (Yan 2021) further improved its performance and generated hydroflattened DEM by adding void filling algorithms to compensate the laser dropouts found at the swath edges and in the close-to-nadir region.

To sum up, state-of-the-arts LiDAR-based water mapping methods exploit the properties of water which are characterized by the low intensity and low point density over water bodies. Also, many studies tried to calibrate the intensity values. However, intensity values are easily confounded by many other factors such as sensor orientation, scan angles, and turbidity and roughness of water surfaces, which require even more concrete calibration than that for the optical images. In the end, LiDAR-based water mapping methods ends up relying on data-driven approaches which require human interventions and high-quality training data; while the performance of data-driven approaches highly depends on the quality of training data and is susceptible to data distribution shifts, just like the case of optical image-based methods.

\subsection{Limitations and opportunities}
A major barrier to developing scalable water body mapping in both optical and LiDAR-based methods has been an inconsistent spectral characteristic of water in remote sensing data (reflectance in optical imagery; intensity in airborne LiDAR). 
This problem becomes more challenging in high-resolution image, it will involve many more small water bodies whose spectral properties are more diverse and more influenced by the surrounding environment. 
% This problem becomes more challenging in high-resolution water body mapping, as higher-resolution images will involve many more small water bodies, and inevitably the spectral properties of water become more diverse and more influenced by the surrounding environment. 
In addition, large-area mapping requires multiple image acquisitions because individual swath width is limited, especially in high spatial resolution imaging; while multiple image acquisitions increase the likelihood of involving different atmospheric conditions or may involve different types of sensors, which will eventually make scalable water mapping much more difficult.

%Intensity is a measure, collected for every point, of the return strength of the laser pulse that generated the point. It is based, in part, on the reflectivity of the object struck by the laser pulse. Reflectivity is a function of the wavelength used, which is most commonly in the near infrared. The strength of the returns varies with the composition of the surface object reflecting the return.

In the meantime, topography is defined by terrain and water, and they interact and determine each other's shape. For the complete topography mapping, both terrain and water maps are needed. In addition, as discussed, we need a three-dimensional terrain to make surface water map truly impactful. This is why airborne laser scanning is necessary for the high-resolution 3D hydrography mapping. National-scale airborne LiDAR data has been widely collected nowadays. It gives us a great opportunity for large-area mapping at high-resolution. However, since the performance of water mapping using LiDAR is not as effective as using optical imagery, airborne LiDAR data has been used only for terrain mapping and has not been effectively used for water mapping.

In this paper, we developed a scalable water mapping workflow that only requires topographic airborne LiDAR data. The workflow uses only the 3D coordinates of the earth and leverages the geometric and signal properties of water from topographic airborne LiDAR data. As the intensity data is not used, any sensor-specific calibration is not needed, and uncertainties from spectral variability can be avoided. Our workflow extracts water segments based on the LiDAR's point density and extends the extent of each segment based on the robust assumption that the local surfacd water is flat. With the proposed method, it is now possible to obtain full 3D topographic map in a fully automated manner.

\section{Proposed water mapping method}
\subsection{Overall strategy}
\label{sec:Finely rasterized DSM generation}
A standard LiDAR sensor for topographic scanning emits near-infrared laser pulses whose most of the energy is absorbed by the water or specularly reflected away from the receiver's field of view. Therefore, water bodies can be characterized by a large number of laser point dropouts in topographic airborne LiDAR data. In other words, local point density of LiDAR point clouds can be a variable that distinguishes water from non-water. However, the tricky part is that even within the same water body, local point densities can vary greatly, and there are non-water regions where the local point density is lower than that of a water body due to occlusion. The main reason for the point density variability within water body is varying incidence angle. Theoretically, the number of recorded laser points should gradually decrease as incidence angle increase. However, not only the surface conditions on the ground and water, but also the sensor orientation due to the physical movement of the airborne platform greatly influences the local point density over the surface. To overcome the limitation of solely relying on point density, previous studies have utilized intensity information. However, intensity values are also greatly affected by sensing conditions, which could make scalable water body mapping more difficult.

The proposed method simply uses only 3D coordinate information of point clouds. It does not used the intensity information. 
%We noted that intensity information would rather be a hindrance to the development of the scalable algorithm. 
Instead, we add an assumption that local surface water is flat. This assumption is robust as the way of water is always downwards and locally connected. 
Figure 1 shows a pictorial example of typical airborne laser scanning. 
\begin{figure}[b]
	\centering
	%\captionsetup{justification=centering}
	\includegraphics[width=0.45\textwidth]{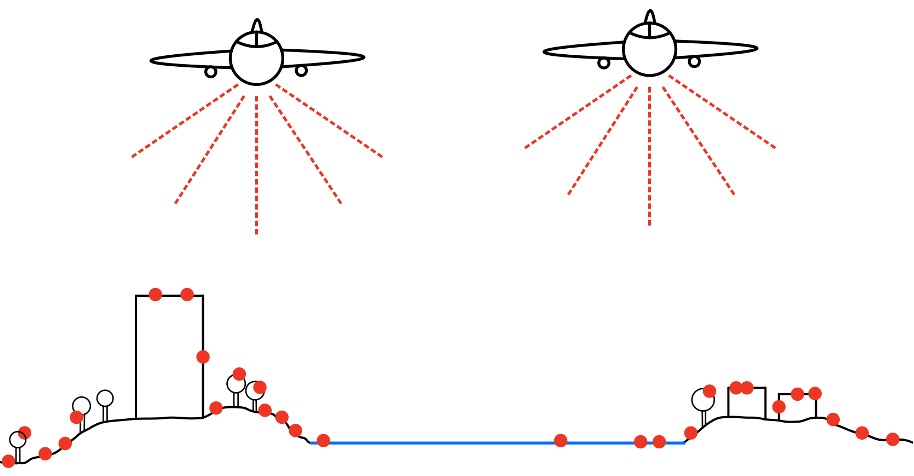}
	\caption{Profile view of an airborne laser scanning}
	\label{fig:fig1}
\end{figure}
The red dot represents points that are reflected and observed by the receiver. Only red dots' 3D coordinates are the input of the proposed method. As shown, low point density can either indicates a water body or an occluded area by buildings. Therefore, the proposed method first considers the area with low point density as the initial water segment but excludes the false water segment caused by occlusion of high-rise buildings. On the other hand, some regions of the water body can have as a high point density as the ground due to the low incidence angle, which makes the initial water segment not comprehensive. Thus, the proposed method extends the initial water segment using the water elevation-based region merging. Specifically, the proposed method extends the initial water segment by merging other water segments whose elevations are the same as the initial water segment and that are connected to the initial segment at the same time. To sum up, the proposed method consists of the following two steps (1) initial water segment extraction, and (2) water elevation-based region merging (WERM).

%Next, the proposed method refines the water body by extending the water b
%Next, considering the elevation of each initial water segment, among terrains with the same elevation, only terrains connected to the initial water segment are remapped to the water body.

%airplane attitude (i.e. pitch) the maximum of the real data records shown in Figure 1(b) lies between 4° and 6° caused by the higher sampling density

% Hence, the point density of lidar data within water areas is usually significantly lower than within land areas.
% but also the systematic changes depending on the angle of incidence, have to be taken into account
% “This can be seen in Figure 1(b) where above incidence angle of 9° hardly any echoes are recorded. Theoretically, under optimal flight and water surface conditions the number of recorded echoes should gradually decrease with incidence angle, but due to rapid changing airplane attitude (i.e. pitch) the maximum of the real data records shown in Figure 1(b) lies between 4° and 6° caused by the higher sampling density in these areas of the selected transect. Within the range of lower incidence angle (<2°) high average intensity and low dropout rate can be found.” Hofle
%Basically, intensity is hard to modelled. To be more accurate, relative orientation/distance/flight atittude/scan angle to the objects, and BRDF should be modelled. But, BRDF is very difficult to obtain. Especially, reflectance of water is quite varying, overlapping strips are difficult to unravel.

\subsection{Initial water segments extraction}

\begin{figure*}
	\centering
	%\captionsetup{justification=centering}
	\includegraphics[width=1\textwidth]{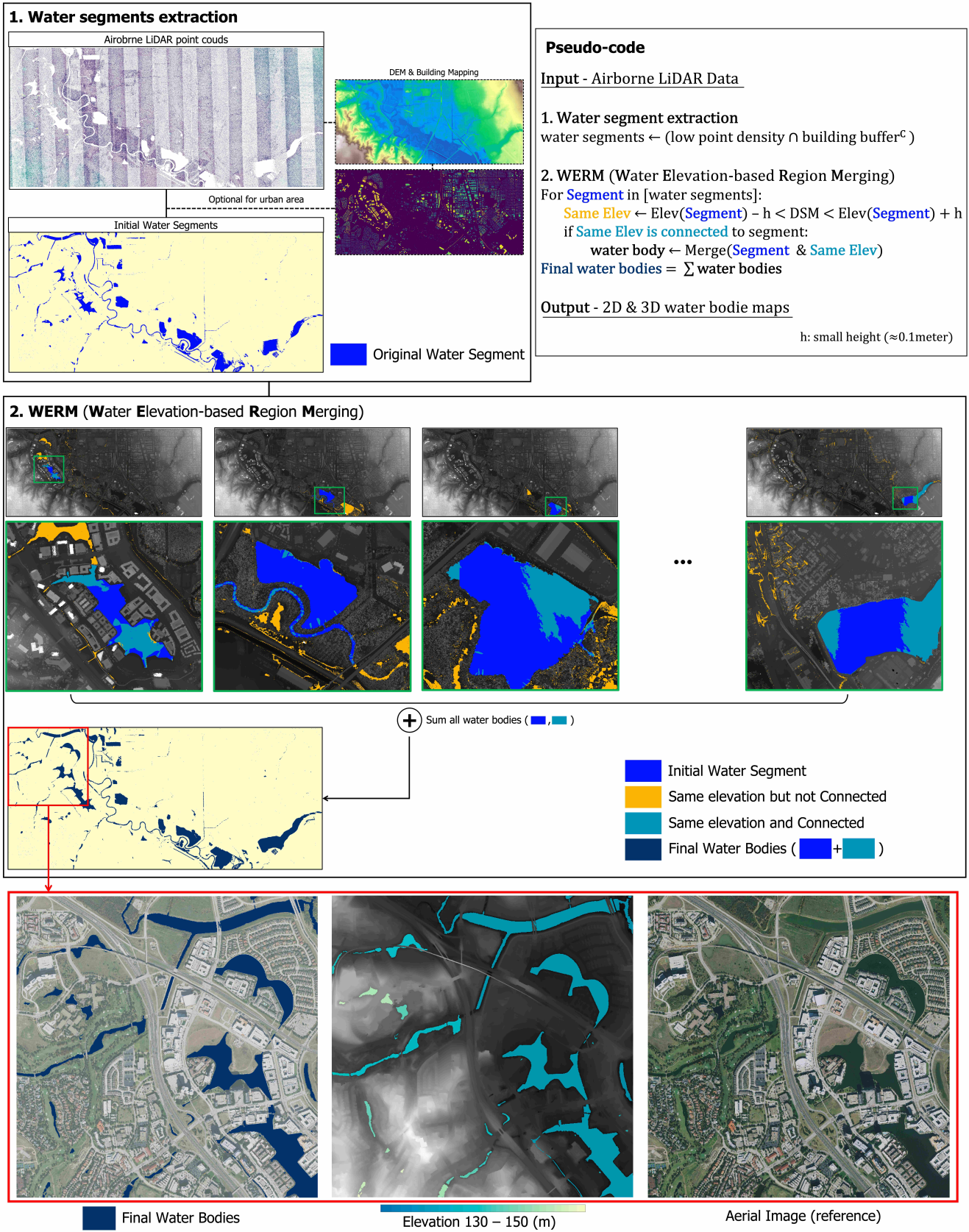}
	\caption{A graphical illustration of the proposed method}
	\label{fig:fig2}
\end{figure*}

In the proposed method, initial water segments are extracted based on the assumption that the point density over water bodies is much lower than in non-water areas as water bodies hardly reflect laser points. Please refer to “1. Water segments extraction” of Figure 2. 

First, the proposed method rasterized the raw airborne LiDAR data into a high-resolution 2D gridded space, digital surface model (DSM). If the point clouds are rasterized into high-resolution 2D gridded space, there will be pixels where any laser point is not registered, and areas with more non-registered pixels are more likely to be water bodies. Our method calculates the average point density ($P$) of a DSM by dividing the number of non-empty cells by the number of total cells. Then, the number of non-empty cells in a certain size of a sliding window ($N pixels$) will have a binomial distribution $B(N, P)$. Specifically, $B(N, P/2)$ was used to compensate for the imbalance of point density due to scanning overlap and to avoid overly detecting water. Based on the binomial distribution, a lower confidence bound was used for the decision boundary of the water classification. Simply put, the center of sliding window will be classified as a water pixel if the sliding window contains relatively fewer numbers of laser points than other windows.

As a default, we rasterized the point clouds into a 0.5-meter resolution grid, and we used a window of 9 x 9 and a confidence level value of 2 (the critical z-score) as a default. In addition, as shown in Figure 1, as lower local point density can be due to an occlusion from adjacent high-rise buildings, the proposed method adopts a 3D building mapping algorithm \citep{song2022towards} as an optional operation to exclude false positives by creating a building buffer. This operation is only for urban areas with high-rise buildings, as non-urban areas rarely have occlusion large enough to be extracted by water.

%Creating a building buffer is an optional process only for urban areas. This is because occluded are by adjacent tall buildings can be extracted as water segment almost occur only in urban areas with many high-rise buildings. This is because large buildings that can cause occlusion of a size that can be extracted as water segment almost occur only in urban areas with many high-rise buildings.

\subsection{Water Elevation-based Region Merging (WERM)}

As the initial water segment is not comprehensive, the proposed method extends the initial water segment based on the assumption that the elevation of the connected water segment is the same. Please refer to “2. Water Elevation-based Region Merging (WERM)” in Figure 2.

First, for each initial water segment, regions of the same elevation are extracted by slicing the DSM based on the elevation. Then, if any extracted region is connected to the initial water segment, the region is merged into the initial water body. In this way, the algorithm WERM expands initial water segments to neighboring regions whose elevations are very close to each initial water segment. Finally, all initial water segments and the merged water segment will be the final surface water map.

Specifically, in WERM, we picked the $10th$ percentile of elevations of each initial water segment as the elevation of the water segment.
% the elevation of each initial water segment was selected by taking the 10th percentile of elevations among each water segment to prevent outliers
As a default, WERM finds regions of the same elevation by slicing the DSM within $\pm 0.1m$ elevation range considering the vertical precision of airborne LiDAR data. Of all initial segments of water, WERM only works for bodies of water larger than 500$m^{2}$ to exclude insignificant water bodies such as small pools and puddles. Lastly, WERM runs twice as the extended water segment after the first run can give a more reliable elevation value of each segment. The default numerical values ($10th$ percentile, $\pm 0.1m$, and 500$m^{2}$) used in WERM are empirically determined and can be modified. A more detailed discussion in parameter values can be found in Section 6.

% Lastly, the elevation of water bodies was selected by taking the 10th percentile of elevations among each water segment to prevent outliers. These parameters were set empirically and found to be robust in diverse topographic airborne laser scanning, but it is worth noting that elevation values cannot guarantee the true elevation as observations of water bodies contain lots of noise. This is because a LiDAR for topographic mapping commonly uses a near-infrared laser, which is absorbed by water and cannot reflect the laser point. Moreover, the elevation of the water bodies is dynamic in nature due to the water cycle. A more detailed description and impacts of water-related parameters are provided in Section 3.2.3.

\section{Datasets and experimental design}
Extensive and various landscapes were used for the evaluation as the development of a scalable water mapping algorithm is our goal. Five different large-scale datasets were employed for the evaluation (Figure 3). The five datasets include three metropolitan areas, one mountainous area, and one coastal area.  Seas, lakes, rivers, streams, ponds, wetlands, ditches, and snows were included in the dataset. Approximately 2,500$km^{2}$ area, 10 billion pixels evaluated at 0.5-meter resolution.

\begin{figure*}
	\centering
	%\captionsetup{justification=centering}
	\includegraphics[width=0.8\textwidth]{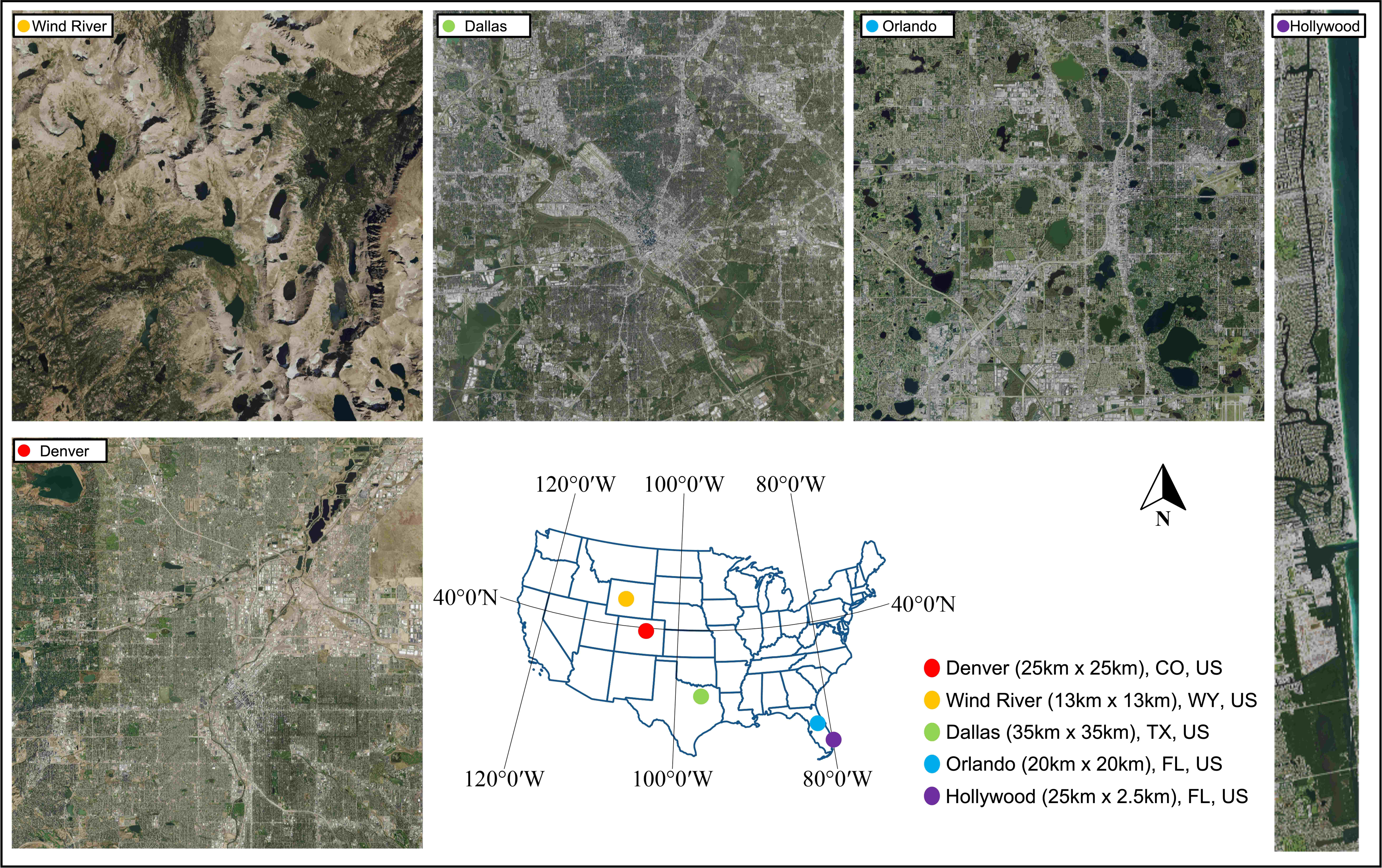}
	\caption{Five large-area datasets over the US: Denver, Wind River, Dallas, Orlando, and Hollywood}
	\label{fig:fig3}
\end{figure*}
%wind river center coordinate, maps: https://www.google.com/maps/place/Wind+River+Range/@42.9215365,-109.3657587,595m/data=!3m1!1e3!4m5!3m4!1s0x8757f965fb70b0a1:0x30f57181f6e643aa!8m2!3d43.1835193!4d-109.6526115
% https://prd-tnm.s3.amazonaws.com/StagedProducts/Elevation/metadata/WY_FEMA_East_2019_D19/WY_FEMA_East_B9_2019/reports/WY_FEMA_East_2019_D19_Lidar_Project_Report_Lot_9_Blocks_9_and_10.pdf

For the quantitative evaluation, we employed the water body layer of the USGS’s High Resolution National Hydrography Dataset Plus (NHDPlus HR) \cite{moore2016road, moore2019user}. NHDPlus HR is an integrated set of geospatial data layers, including the best available National Hydrography Dataset (NHD) and the National Watershed Boundary Dataset (WBD). Also, we compared the proposed method to NDWI-based thresholding methods.

Specifically, we used two different NDWI-based methods that have different levels of privileges. For the first method, we found a single optimal threshold value that makes the method obtain the highest possible overall accuracy compared to the ground truth (NHDPlus HR) for each experimental area. For the second method, we divided each dataset into small-sized tiles and found the optimal threshold value for every individual small-sized tile. Thus, the first method (“NDWI-G-Opt”) creates a surface water map using the global optimal threshold value; while the second method (“NDWI-L-Opt”) creates a surface water map that is stitched with small-sized tiles that are locally optimized to obtain the highest performance for each tile. Although the two NDVI-based methods are not feasible in real practice as the groundtruth must be unknown, we used them to evaluate the proposed method’s performance with more competitive criteria. On the other hand, for the proposed method, we used the same parameter values for the entire dataset. No site-specific optimization was carried out. In this way, we investigate the scalability and the performance of the proposed method in comparison with highly competitive NDVI-based methods to which we have given special benefits.

\begin{table*}[width=2\linewidth]%[width=.9\linewidth,cols=4,pos=h]
\caption{Summary of the experimental datasets}\label{tbl1}
\begin{tabular*}{\tblwidth}{LLLLLL}
\toprule
Name& Denver& Wind River& Dallas & Orlando & Hollywood\\
\midrule
Location& Denver, CO & Wind River, WY & Dallas, TX & Orlando, FL & Hollywood, FL\\
\midrule
Dimension & 25-km by 25-km & 13-km by 13-km & 35-km by 35-km & 20-km by 20-km &  25-km by 2.5-km\\
\midrule
Geography & Metropolitan area & Mountainous area & Metropolitan area & Metropolitan area & Coastal area\\
\hline
& Leica TerrianMapper & Leica TerrianMapper & Leica ALS 80 & Leica ALS 80 & Riegl VG-1560i\\
& 2.4 points$/m^2$ & 2.0 points$/m^2$ & 3.0 points$/m^2$ & 9.8 points$/m^2$ & 8.2 points$/m^2$\\
Scanning details & AGL: 2.7-3.1km & AGL: 3.0-3.2km & AGL: 1.8km & AGL: 1.4km & AGL: 1.3km\\
& scan angle: 40$^{\circ}$ & scan angle: 40$^{\circ}$ & scan angle: 35$^{\circ}$ & scan angle: 40$^{\circ}$ & scan angle: 60$^{\circ}$ \\
& overlap: 20\% & overlap: 25\% & overlap: 30\% & overlap: 60\% & overlap: 30\%\\
\hline
Terrain elevation & 25-km by 25-km & 13-km by 13-km & 35-km by 35-km & 20-km by 20-km &  25-km by 2.5-km\\
\midrule
LiDAR acquisition
&
2020.05.-2020.06.
&
2019.08.-2019.09.
&
2019.03.-2019.07.
&
2018.12.-2019.03.
&
2018.06.
\\
\midrule
NAIP acquisition
&
2019.08.-2019.09.
&
2019.08.%2019.08.27
&
2020.10.-2020.11.
&
2019.11.
&
2019.11.
\\

\bottomrule
\end{tabular*}
\end{table*}

Table 1 summarizes the five datasets. Not only geographic diversity but also diversity of laser scanning conditions are taken into account for the assessment. All airborne LiDAR data were from the USGS's 3D Elevation Program (3DEP) (https://www.usgs.gov/3d-elevation-program)\cite{stoker2022accuracy}. For the NDWI-based water mapping, near infrared and green bands of the National Agriculture Imagery Program’s optical imagery (NAIP) was used.

% \begin{table*}%[width=.9\linewidth,cols=4,pos=h]
% \caption{This is a test caption. This is a test caption. This is a test
% caption. This is a test caption.}\label{tbl1}
% \begin{tabular*}{\tblwidth}{@{} LLLLLLL@{} }
% \hline
% \multirow{2}{*}{Dataset} &
% \multicolumn{1}{c}{A} &
% \multicolumn{4}{c}{B} &
% \multicolumn{1}{c}{C} \\
% & O.B.R & A.R & O.B.R & A.R & O.B.R & A.R \\
% \hline
% D1 & 2.1\% & 2.1\% & 2.1\% & 2.1\% & 2.1\% & 2.1\% \\
% \hline
% D2 & 11.6\% & 11.6\% & 11.6\% & 11.6\% & 11.6\% & 11.6\% \\
% \hline
% D3 & 5.5\% & 5.5\% & 5.5\% & 5.5\% & 5.5\% & 5.5\% \\
% \hline
% \end{tabular*}
% \end{table*}

The experimental dataset is very large. Providing averaged quantitative results for all datasets might be not sufficient to investigate the performance in-depth. Also, NHD plus datasets include snows, wetlands, and some mangroves as water class while the proposed method and NDWI-based method are intended to extract only surface water bodies. % airborne optical and LiDAR observations are not appropriate for extracting

To address these issues, tile-based evaluation was introduced in addition to the conventional quantitative evaluation to conduct more detailed evaluations in various ways. The tile-based evaluation method proceeds as follows. First, the entire experimental area was divided into small-sized tiles. Then, quantitative evaluation was conducted after excluding tiles that have significant amounts of wetlands and snow cover. In addition, we performed a separate evaluation for only selected challenging tiles based on the magnitude of disagreements among surface water maps. To be specific, we calculated a tile-based difference in intersection over unions (IoU) of the proposed method and the NDWI-based method. Then, we selected challenging areas based on the tile-based IoU difference. Lastly, we selected three sets of tiles for each dataset and investigated the difference both quantitatively and qualitatively. In this way, we tried to evaluate the performance of the proposed algorithm effectively and objectively in extensive datasets.

\section{Experimental results}
\label{sec:Experimental results}

\begin{figure*}
	\centering
	%\captionsetup{justification=centering}
	\includegraphics[width=1\textwidth]{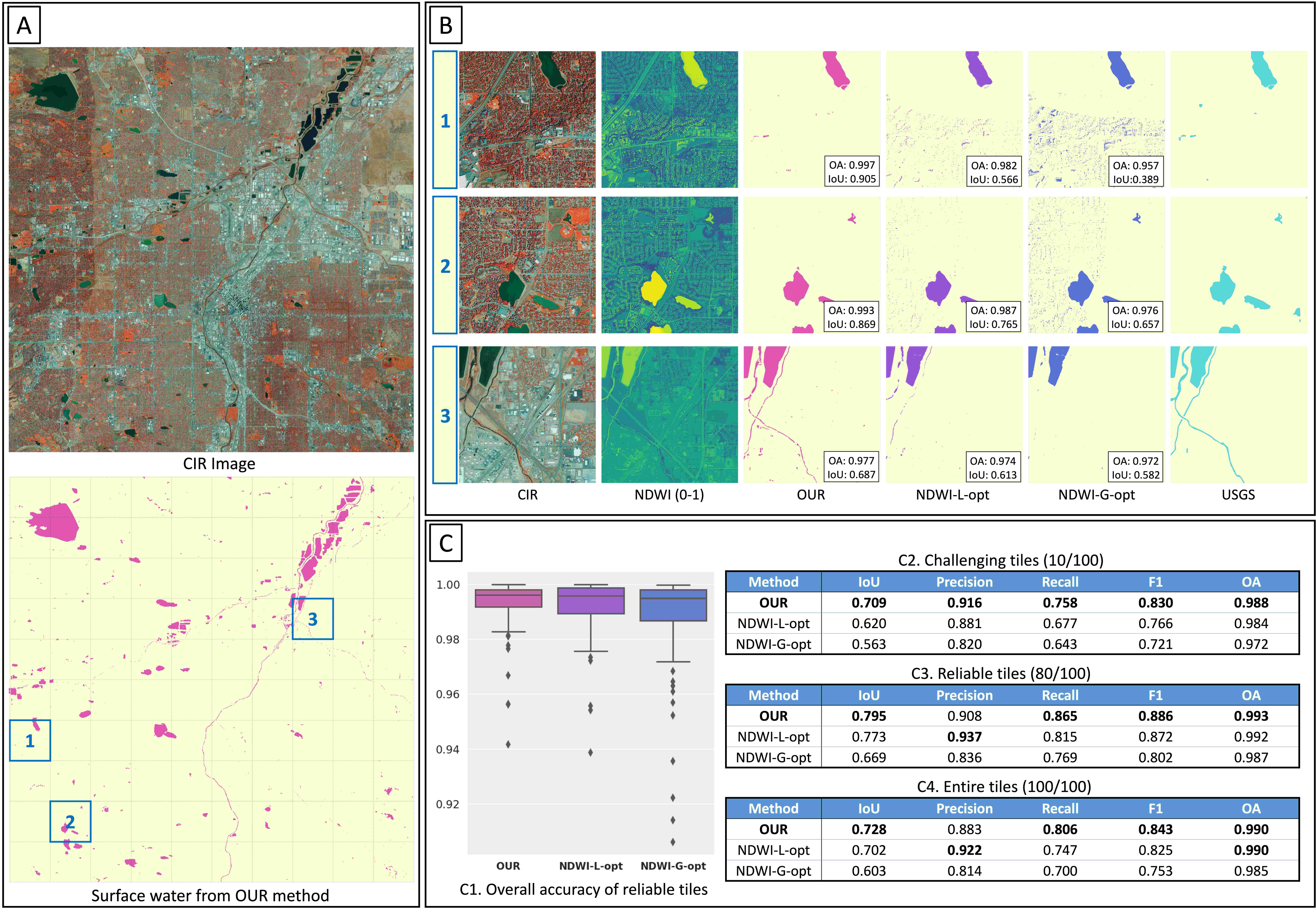}
	\caption{Results of the Denver dataset}
	\label{fig:fig4}
\end{figure*}

\begin{figure*}
	\centering
	%\captionsetup{justification=centering}
	\includegraphics[width=1\textwidth]{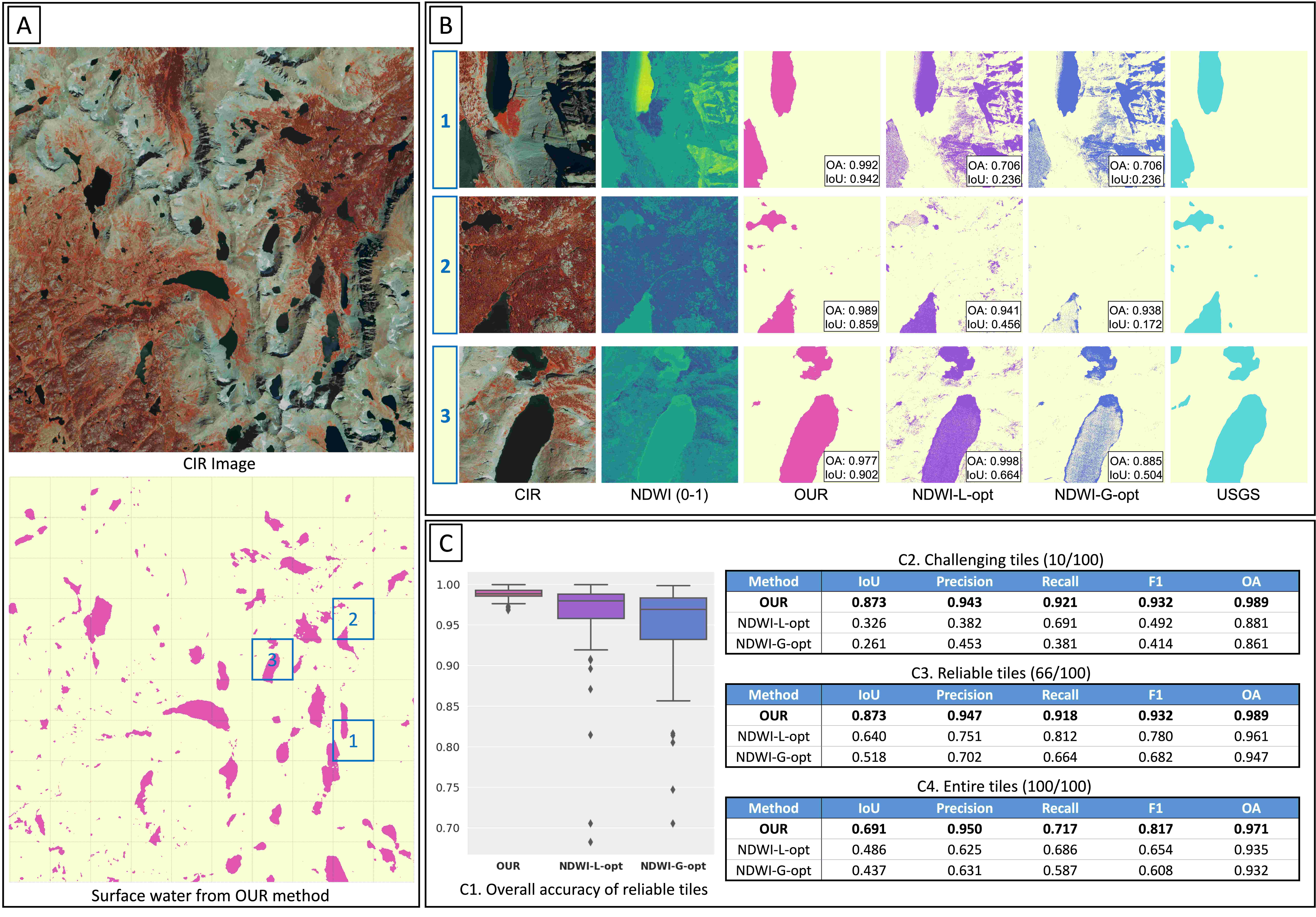}
	\caption{Results of the Wind River dataset}
	\label{fig:fig5}
\end{figure*}

\begin{figure*}
	\centering
	%\captionsetup{justification=centering}
	\includegraphics[width=1\textwidth]{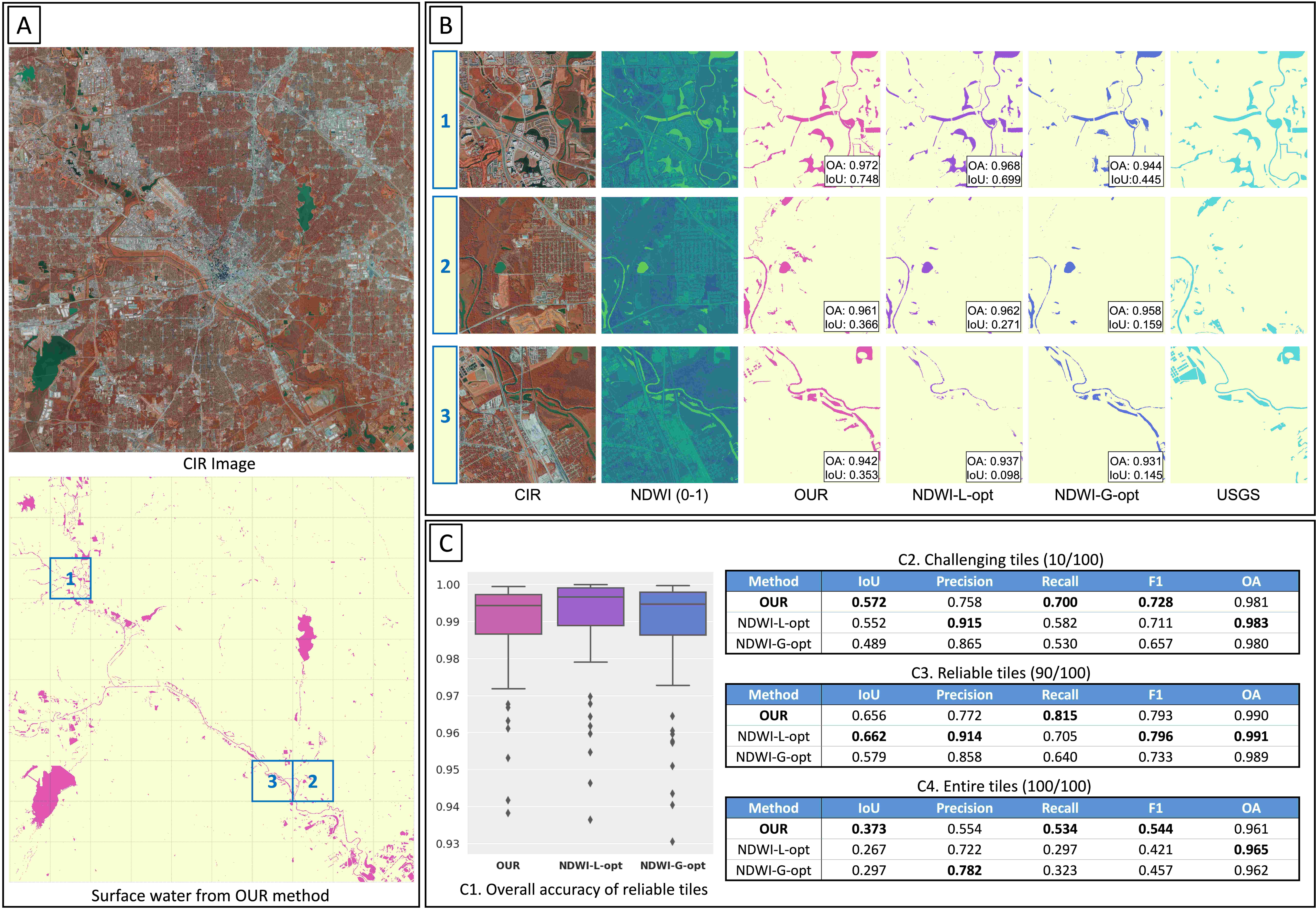}
	\caption{Results of the Dallas dataset}
	\label{fig:fig6}
\end{figure*}

\begin{figure*}
	\centering
	%\captionsetup{justification=centering}
	\includegraphics[width=1\textwidth]{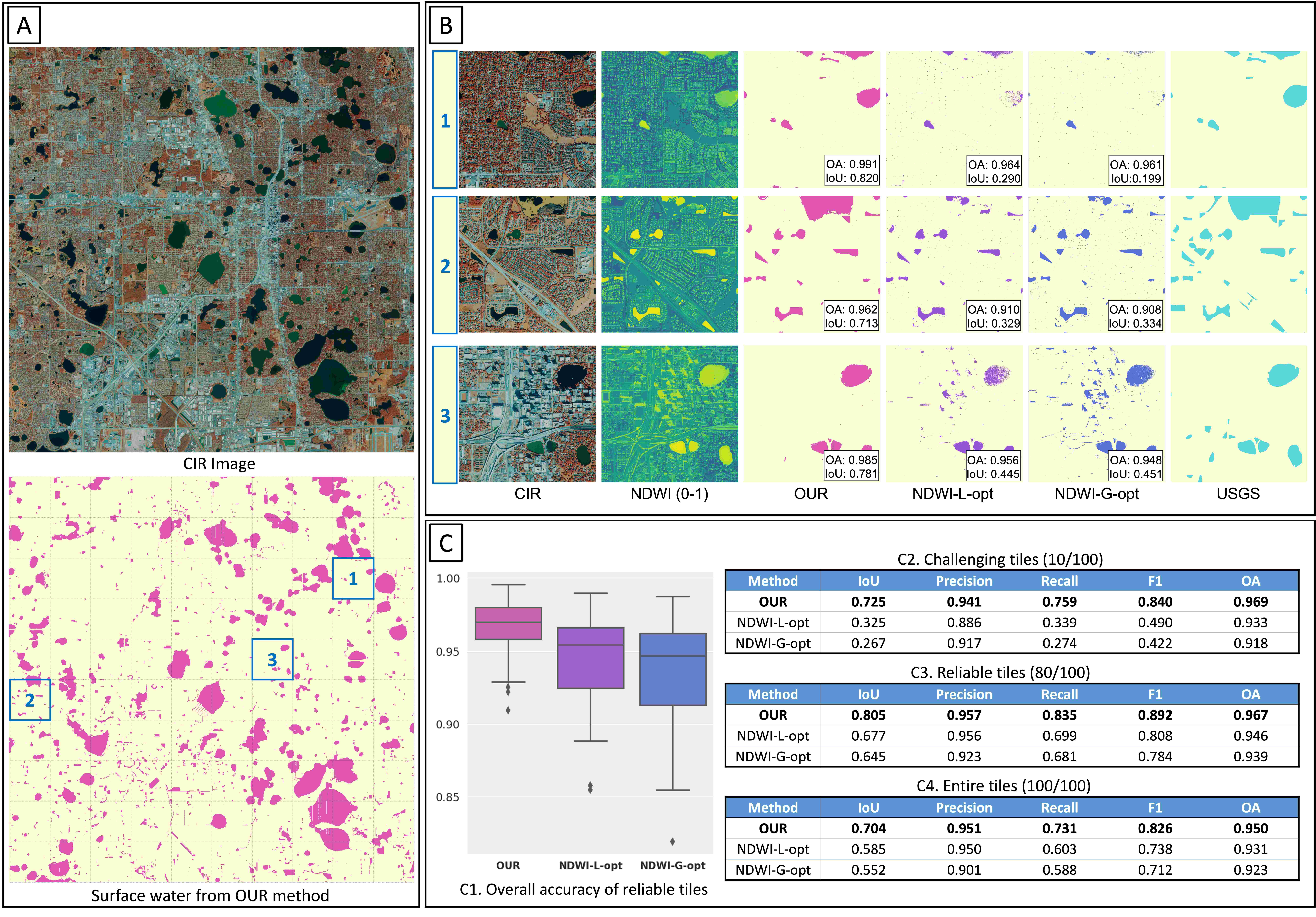}
	\caption{Results of the Orlando dataset}
	\label{fig:fig7}
\end{figure*}

\begin{figure*}
	\centering
	%\captionsetup{justification=centering}
	\includegraphics[width=1\textwidth]{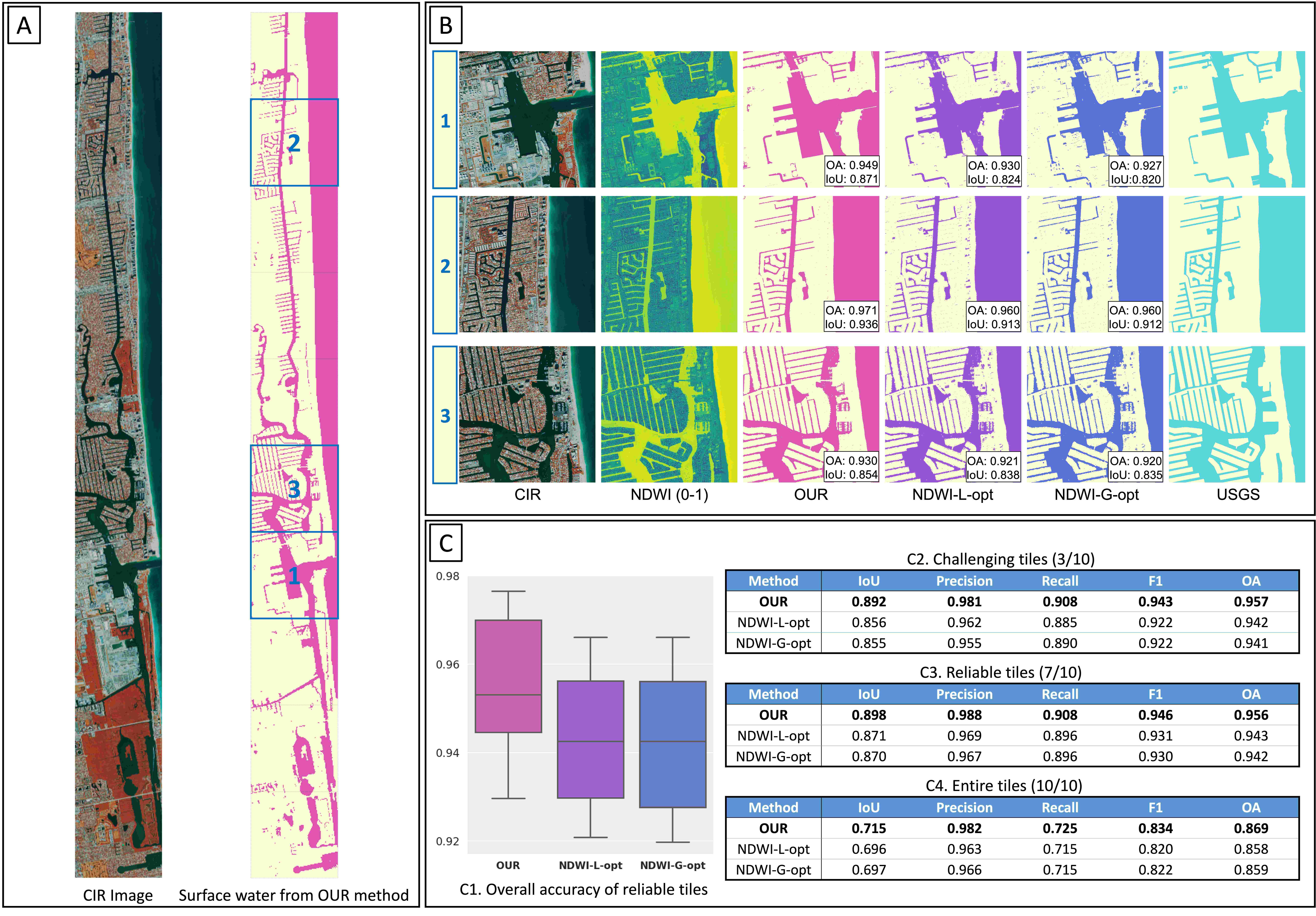}
	\caption{Results of the Hollywood dataset}
	\label{fig:fig8}
\end{figure*}

Figure 4 shows the results for the Denver dataset. Figure 4-A shows the CIR image and extracted surface water map from the proposed method for the entire Denver dataset. OUR represents surface water maps generated from the proposed method. Figure 4-B highlighted three subset regions that showed significantly different results. CIR, NDWI, and four different surface water maps were visualized. Overall accuracy (OA) and IoU compared to USGS's surface water map were also noted. Figure 4-C shows the statistics of the three methods. C1 boxplot displays the overall accuracies of the three methods. C2, C3, and C4 show the comparison results for three different sampled sets. C2 shows the results from the top 10 challenging tiles that showed the largest IoU difference among different methods. C3 shows the results after excluding tiles whose groundtruths contain significant amounts of snow or wetland cover. Lastly, C4 shows the results from the entire tiles of the Denver dataset.

As shown in Figure 4-A, it is confirmed that the visually identifiable large water bodies are generally well-extracted. The entire 25-km by 25-km test area was divided into 100 square tiles, of which the tile showing the largest difference were enlarged and displayed in Figure 4-B. As shown in the first row of Figure 4-B, the proposed method hardly generated false positives in the urban area compared to the other two NDWI-based methods. NDWI-L-opt used the locally optimized threshold to obtain the highest possible OA in the given tile. Even though it is tuned using the USGS's water map, there are a large amount of spurious false positives in the urban area. NDWI-G-opt was able to detect some water bodies that were not extracted in NDWI-L-opt, but more false positives were detected in the urban area, resulting in lower OA and IoU.

Similarly, in the second row of Figure 4-B, the proposed method accurately extracted water bodies. However, NDWI-L-opt failed to detect some water bodies; while NDWI-G-opt generated too many false positives. Lastly, in the third row, the proposed method showed very similar results to the USGS water body. A closer look at the stream in the upper left shows that the proposed method can depict the detailed branches of the stream in more detail than the body of water in the USGS. On the other hand, the two NDWI-based methods were not able to accurately detect streams. If the NDWI threshold had been set to a lower value, the NDWI-based method would have been able to detect the stream, but false positives would have increased significantly.

As shown in boxplot C1 in Figure 4, overall accuracies are not significantly different although the proposed method has the least deviation overall. C2 shows the quantitative result by selecting only the top 10 tiles with the largest difference in IoU value among all 100 tiles. The proposed method recorded the highest accuracy in all metrics. The recall was particularly higher than other methods, indicating that the proposed method can achieve a high detection rate in small water bodies such as streams, as confirmed in the third row of Figure 4-B. The following C3 is the evaluation result using the filtered groundtruth, excluding 20 tiles that have a fairly large error in the water map of the USGS. Lastly, C4 is the evaluation result using the entire 100 tiles without filtering any tiles. In both C3 and C4, the proposed method recorded higher accuracies than NDWI-based methods overall. NDWI-L-opt obtained the highest precision. This is partly because NDWI-L-opt is tuned to get the highest OA for each tile, which makes NDWI-L-opt tend to map water bodies passively and avoid false positives.

% Wind River

Figure 5 shows the experimental results of the Wind River dataset. Similar to the layout in Figure 4, Figure 5-A shows the CIR image and the surface water map generated by the proposed method for the entire area, Figure 5-B shows three subset regions, and Figure 5-C shows the statistics of the experiment results.

As can be seen in Figure 5-A, the Wind River dataset is a completely mountainous terrain with no man-made structures, and steep and rugged terrains cast many shadows. Seeing the surface water map generated by the proposed method, it can be confirmed that the proposed method distinguishes between water areas and shadow areas that are difficult to distinguish visually.

Figure 5-B selects and shows three areas where there is a large difference in accuracy between water extraction methods. As shown in the first row, the proposed method clearly distinguishes the water body and the shadow caused by the terrain and records very high accuracy, but the method using the NDWI fails to distinguish between water and shadow despite applying an optimal threshold. In the NDWI-based methods, there were noises in the areas shaded by rugged terrain and trees as shown in the second and third rows. On the other hand, the proposed method generally showed very similar results to USGS and recorded very high accuracy. One error in the upper left corner of the third row was due to the snow cover.

As shown in boxplot C1 in Figure 5, the proposed method recorded higher OAs in general, and its variation was also very low. On the other hand, the NDWI-based method recorded significantly lower OA. Moreover, for all quantitative metrics in C2–C4, our proposed algorithm recorded significantly higher accuracy than NDWI-based methods.

% Dallas
The Dallas dataset is a 35-km by 35-km metropolitan area covering various types of water bodies. Including several large lakes, the Trinity River and streams branching off from it are widely distributed. There are also many small wetlands, streams, and ponds throughout.

Figure 6-B shows three sampled areas that show significant differences among surface water maps. The first row shows the reservoirs, streams, and lakes of the urban area. The proposed method recorded the highest values in both OA and IoU, and when visually compared with the USGS water body, the water body layer generated by the proposed method showed little error. Rather, small creeks that were not extracted in USGS were well detected by the proposed method. Meanwhile, NDWI-L-opt and NDWI-G-opt were not able to extract water bodies as well as the proposed method. In particular, NDWI-G-opt had lots of omission errors. Considering that NDWI-G-opt used a single threshold for the entire area to yield the highest OA, the result indicates that a single fixed threshold for large areas cannot guarantee satisfactory performance. In the second row of Figure 6-B, NDWI-L-opt recorded the highest overall accuracy. However, it was found that the USGS water layer has some errors probably due to the temporal discrepancy. When compared to the CIR image, the proposed method extracted small creeks well; while the two NDWI-based methods failed to detect creeks successfully. This sample shows an example that differences in quantitative metrics do not accurately represent actual performance. A similar result was also confirmed in the third row of Figure 6-B. %A close comparison with CIR and NDWI showed that the water body generated by the proposed method showed higher accuracy than the USGS water layer as well as the other two NDWI-based methods.

As shown in Figure 6-C, the overall performance of the proposed method was quite similar to that of NDWI-L-opt. In general, the proposed method tends to detect more water, considering the high recall and low precision. This might be attributed to the fact that the proposed method detects small water bodies more comprehensively as shown in Figure 6-B. Considering that NDWI-based methods find the optimal threshold using groundtruth, the result confirms the proposed method could be a better alternative way for surface water mapping.

% Orlando

The Orlando dataset covers a 20-km by 20-km metropolitan area. The area is dotted with hundreds of lakes and ponds over very flat terrain. As shown in Figure 7-A, lakes and ponds show various optical characteristics, which makes it challenging for image-based methods to extract accurate water bodies. However, since the proposed method exploits laser property and operates with the robust assumption that the connected water has the same elevation, it successfully extracted lakes and ponds.

Figure 7-B highlighted the three sampled regions. The first row contains several small lakes and ponds in the urban area. The proposed method showed satisfactory results when compared with CIR images and USGS's water layer, and recorded high accuracy in OA and IoU. Meanwhile, two NDWI-based methods failed to successfully extract some lakes and ponds. %This is because there are many areas with higher NDWI values than water, so lowering the threshold to sample all bodies of water will cause too many false alarms. 
This is because some non-water areas have higher NDWI values than water bodies. Lowering the threshold would have allowed all the water bodies to be extracted, but that would have generated too many false positives. NDWI-L-opt detected a larger amount of water as it used a lower threshold than NDWI-G-opt, but it costs more noise in non-water areas. The second row shows a unique characteristic of the proposed method in an algae-covered lake.
Unlike NDWI-based methods, the proposed method extracted the complete water body even where algae were covered. This is because algae-covered lakes still have the same level as other algae-free portions.
% algae good, mangrove bad

The third row shows the results of tiles where high-rise buildings and lakes coexist. As shown in the NDWI image, the shadows caused by high-rise buildings sometimes have higher NDWI values than the water body, resulting in many false positives in NDWI-based methods. On the other hand, since the proposed method uses a building buffer, most of the false positives could be prevented.

The proposed method also outperformed NDWI-based methods in quantitative metrics as shown in Figure 7-C. In challenging tiles, the performance gap was distinctive, especially in IoU, recall, and F1. This is because NDWI-based methods kept their thresholds low to retain overall accuracy high. A similar pattern is shown in C3 and C4 as well.

% Hollywood

The Hollywood dataset covers a coastal urban area facing the ocean and is composed of various shapes of creeks, bays, and artificial waterways. As shown in Figure 8-A, the proposed method extracted water bodies well overall.

Figure 8-B shows a total of 3 tiles out of 10 tiles in the Hollywood dataset. Overall, both the proposed method and the two NDWI-based methods showed satisfactory performance. In all cases, OA and IoU were higher than 0.9 and 0.8, respectively. Compared to the USGS water body layer, the main errors were associated with the shadow that created by tall buildings along the coast. Otherwise, errors were either small-sized, insignificant inland water bodies or false negatives due to water transport vehicles parked along the waterway. Among different methods, the proposed method obtained the highest accuracy in both OA and IoU as it hardly generates errors from shadows or occlusions. One unique error observed in this dataset was a false negative due to the rapid current from the boat in the middle of a large water body. This is because the spectral characteristics of the water changed temporarily due to the rapids. Although such a small size error can be easily removed by post-processing, this case confirms the advantage of the proposed method that it is hardly affected by water surface conditions.

Figure 8-C provides quantitative results. Three excluded tiles of C3 are the bottom 3 tiles as the USGS's water map classified mangroves as water bodies. Except for those three tiles, all methods showed satisfactory results, and the proposed method obtained the highest accuracy in all metrics in all cases.

\section{Discussion}
\subsection{The impact of parameters}

\begin{figure*}
	\centering
	%\captionsetup{justification=centering}
	\includegraphics[width=1\textwidth]{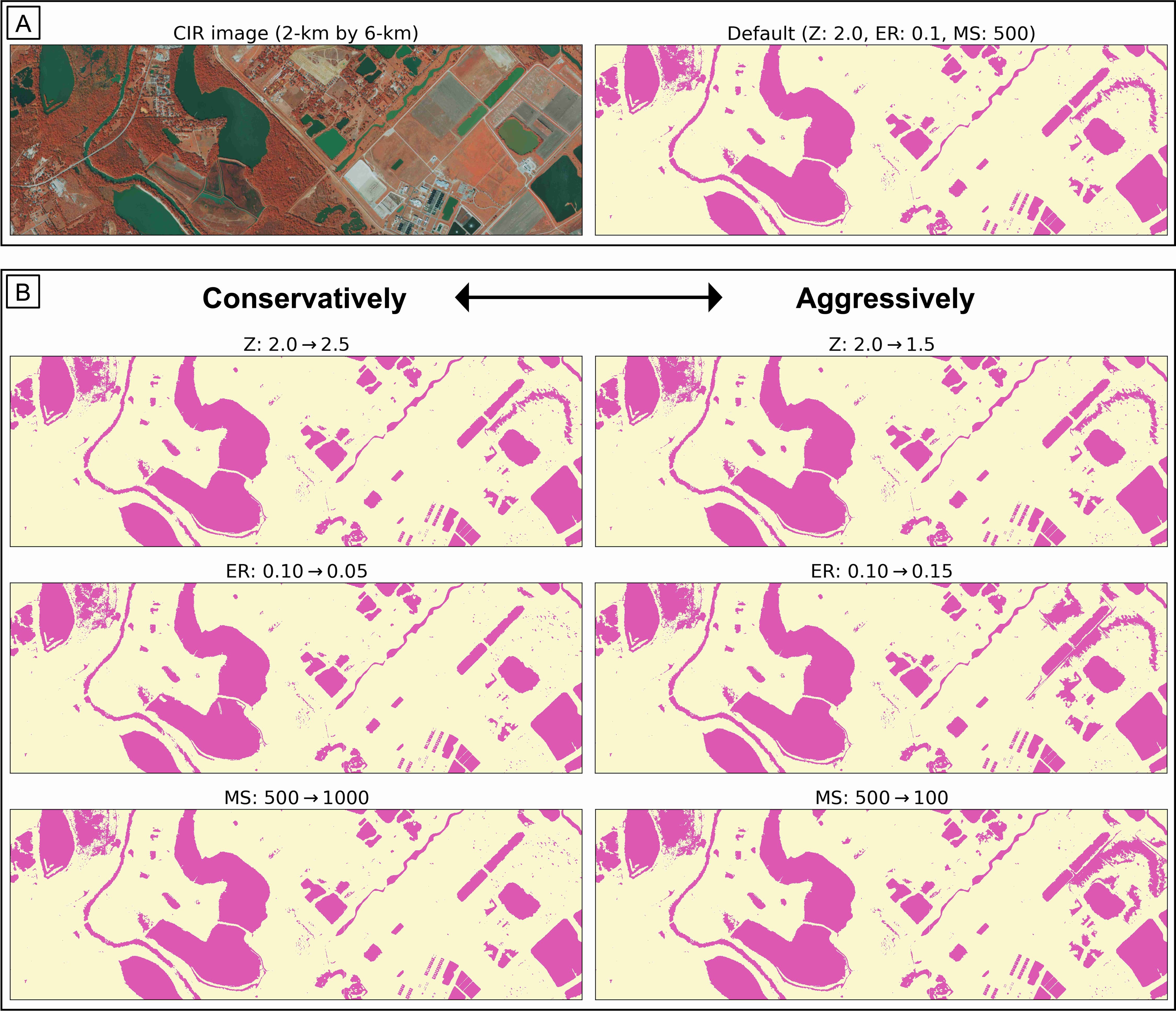}
	\caption{The impact of parameters: z-score (Z), elevation range (ER), and minimum size (MS)}
	\label{fig:fig9}
\end{figure*}

As confirmed by experimental results with extensive datasets with different laser scanning conditions, the proposed method can generate highly accurate surface water maps without additional parameter tuning. Additional experiments found that tuning the parameters within reasonable ranges had little effect on the overall statistics. Nevertheless, we have found that additional parameter tuning sometimes can produce more accurate results.

In this section, we provided experimental results after changing three key parameters: z-score (Z), elevational range (ER), and minimum size (MS). Z determines the confidence interval for finding the initial water segment. For example, a Z of 1.96 represents a 95\% confidence level from a standard normal distribution. The default value for Z is 2 in our method. We conducted additional experiments by changing Z to either 1.5 or 2.5. Thus, if Z is 1.5, it is expected that the method will detect water more sensitively. When Z is 2.5, the method will detect water more conservatively. ER is a small elevation interval for finding areas with the same elevation in the algorithm WERM. The default of ER is 0.1 (a range of $\pm$ 0.1m) considering the typical vertical precision of airborne laser scanning systems. We altered the default value to either 0.05 or 0.15 so that each case extends its initial water segment more conservatively and more aggressively. Finally, the parameter MS determines the minimum size of the initial water segment to be extended to the adjacent area. To prevent insignificant bodies of water, such as puddles, from extending to the surrounding terrain, the default is set to 500 square meters. In this parameter analysis, we experimented with 100$m^{2}$ and 1000$m^{2}$, making the algorithm expand the initial water segment more conservatively and aggressively, respectively.

Among the entire experimental areas, we excerpted one area where the difference according to the parameters was the most prominent. 
%Figure 9 shows the extracted surface water map with different parameter values. 
Figure 9-A shows the CIR image and the surface water map generated from the proposed method with the default parameter. This area is a part of the Dallas dataset. It has the Trinity River meandering on the left and the Dallas Southside Wastewater Treatment Plant on the right. It is a very flat terrain, with a variety of water bodies, including rivers, wetlands, lakes, reservoirs, water tanks, ponds, and puddles.

Figure 9-B shows the resulting surface water map after varying three parameters' values (i.e., Z, ER, and MS). On the left side of the center, parameters are adjusted so that the water body can be extracted more conservatively, and on the right side, parameters are adjusted so that the water body can be extracted more aggressively. First, when Z was changed from the default value to 2.5 and 1.5, respectively, no significant changes occurred overall. Considering that Z works as a reference point for distinguishing water and non-water pixels using local point density, results suggest that the classification of water pixels based on the point density is quite robust. On the other hand, when the ER was changed from the default value to 0.05 and 0.15, respectively, significant changes were observed in the resulting surface water maps. First, when ER was changed to 0.05, the error as if the small pond was flooded to the surrounding terrain on the right side of Figure 9-A was improved. On the other hand, when the ER is changed to 0.15, this "flooded" error becomes wider. A similar pattern was observed when the parameter MS was changed. When MS was set to 1000, most small ponds were not extended as they were smaller than 1,000$m^{2}$. However, when is set to 100, some smaller ponds were extended to adjacent terrains.

Figure 9 shows an example of how parameters can affect the resulting surface water map. The results showed that ER and MS parameters can significantly affect the resulting surface water map. However, since the two parameters are physically meaningful parameters, as they describe water level and water area, respectively, the parameter tuning and its result are quite intuitive and easy to explain. In other words, the error and uncertainty of the proposed method can be managed relatively well compared to that of the optical image-based method. %Finally, the authors would like to mention that this example was one of the most drastically changed areas after changing parameters.

%%%%%%%%%%%%%%%%%%%%%%%%%%%%%%%%%%%%%%%%%%%%%%%%%%%%%%%%%%%%%%%%%%%%%%%%%%%%%%%%%%%%

\subsection{3D surface water map}

The proposed method can generate a 3D surface water map that includes elevation information of each water body as a by-product. 2D surface water map is very useful geospatial data to support numerous studies and decision makings. However, a 2D surface water map alone cannot say where water is flowing from and where it goes. In addition, 3D information of water is helpful to monitor the water cycle or to monitor and prevent water-related pollution and disasters. Until now, most 3D hydrology has been produced by placing a 2D water layer created from optical images on a DEM. For example, the USGS commissions and distributes the production of hydro-flattened DEMs, which requires post-processing of DEM using a separate GIS layer of water. This procedure necessarily entails additional errors due to temporal difference and registration errors.
%However, as this procedure requires the fusion of different datasets from totally different sensors, it necessarily entails more errors including registration errors. 
With the proposed method, a full 3D topography (both terrain and water) can be produced with the same original data \citep{song2022new}. 
%However, since the proposed method allow entire 3D topography, both terrain and water, to be produced with the same original data, complicated procedures and errors can be avoided. 
We expect the proposed method can shorten complicated procedures for hydro-flattened DEM and 3D water mapping while improving their accuracies
%his not only improves the accuracy of the 3D water map but also improves the accuracy of the DEM.

\section{Conclusion}

This paper presents a surface water mapping method that provides reliable and accurate results at high resolution across a wide variety of landscapes. The proposed method uses only airborne LiDAR data based on the natural law that surface water is flat as gravity always pulls liquid molecules down. Avoiding the use of optical properties, our method enables fully automated and scalable surface water mapping. Extensive experiments with large ($\approx$ 2,500$km^{2}$) and diverse landscapes (urban, coastal, and mountainous areas) confirmed that our method can generate more accurate results than NDVI-based methods even without any parameter tunings. We confirmed its overwhelming performance of water extraction in varied conditions placed in diverse environments. We confirmed its advantage, especially in steep and rugged mountains with shadows and perennial snows and complex urban environments with very small water bodies. We expect the proposed method will expedite an accurate large-scale full 3D topography mapping.

% \bibliographystyle{elsarticle-num-names}
% \bibliographystyle{cas-model2-names}

\bibliography{cas-refs}

\end{document}